\pdfoutput=1

\documentclass[11pt]{article}

\usepackage[final]{acl}

\usepackage{amsfonts}
\usepackage{amsmath}
\usepackage{booktabs} 
\usepackage{tabularx}

\usepackage{times}
\usepackage{latexsym}

\usepackage[T1]{fontenc}

\usepackage[utf8]{inputenc}

\usepackage{microtype}

\usepackage{inconsolata}

\usepackage{graphicx}

\title{Using Text-Based Causal Inference to Disentangle\\ Factors Influencing Online Review Ratings}

\author{Linsen Li \and Aron Culotta \and Nicholas Mattei \\
        Department of Computer Science\\
        Tulane University, New Orleans, LA, USA}

\begin{document}
\maketitle

\vspace{-.5cm}

\begin{abstract}
Online reviews provide valuable insights into the perceived quality of facets of a product or service. While aspect-based sentiment analysis has focused on extracting these facets from reviews, there is less work understanding the impact of each aspect on overall perception. This is particularly challenging given correlations among aspects, making it difficult to isolate the effects of each. This paper introduces a methodology based on recent advances in text-based causal analysis, specifically CausalBERT, to disentangle the effect of each factor on overall review ratings. We enhance CausalBERT with three key improvements: temperature scaling for better calibrated treatment assignment estimates; hyperparameter optimization to reduce confound overadjustment; and interpretability methods to characterize discovered confounds. In this work, we treat the textual mentions in reviews as proxies for real-world attributes. We validate our approach on real and semi-synthetic data from over 600K reviews of U.S. K-12 schools. We find that the proposed enhancements result in more reliable estimates, and that perception of school administration and performance on benchmarks are significant drivers of overall school ratings.

\end{abstract}

\section{Introduction}
Understanding the influence of specific aspects mentioned in text reviews on the overall ratings of products or services is a complex yet important endeavor in many industries.
For example, in education, how does feedback on academic performance or facility quality influence a school’s overall ratings? Precisely quantifying this influence can help businesses identify key areas for improvement.

Traditional approaches to this problem include aspect-based sentiment analysis, which extracts sentiments tied to predefined aspects \cite{zhang2022survey, kandhro2024exploring}, and exploratory analyses that measure the correlation between certain terms and overall rating~\cite{geetha2017relationship}. However, these methods generally fail to account for confounding variables, leading to biased results. For example, consider a school often praised for its academic performance, with reviews about its excellent programs and great teachers. Traditional analysis might directly link these positive attributes to the school's high ratings. However, if these reviews also frequently mention extensive extracurricular opportunities or high parental involvement, it could imply that the school's high ratings reflect its socioeconomic advantages, not educational quality alone. 
Overlooking such factors could lead to incorrect assessments of the true influence of educational quality on overall perceptions.

Our goal is to estimate the impact of aspects mentioned in reviews on overall ratings. To achieve this, we have developed a causal inference framework to control for confounding variables within text. Our framework starts by identifying an aspect of interest from the text related to an entity, e.g., if the school reviews frequently praise facility quality. We then utilize the remaining text of the entity, excluding the aspect-related text, as covariates to analyze the overall rating for the entity. Here, the textual content is treated as a proxy for real-world factors that may influence an entity’s rating. This approach helps us control for confounders associated with the entity and isolate the treatment effect of the specific aspect on the overall rating.

\vspace{-.2cm}
\paragraph{Contribution}

We apply CausalBERT~\cite{veitch2020adapting} to estimate the effects of specific topics on overall review ratings, isolating genuine influences from other textual elements.  We enhance CausalBERT by (i) integrating Temperature Scaling to calibrate propensity scores; (ii) optimizing a key hyperparameter that balances treatment and outcome prediction, reducing overadjustment for confounds; (iii) employing interpretability methods to characterize discovered confounds. We validate our approach using 600K U.S. K-12 school reviews from GreatSchools.org, finding that issues of administration personnel and academic performance are significant drivers of perceived school quality.

\section{Methods}
\label{sec:methods}
\label{subsec:estimation-treatment-effects}
We apply the potential outcomes framework \cite{neyman1923application}, observing for each subject (school) \(i\) a tuple \((X_i, Y_i, T_i)\), where \(X_i \in \mathbb{R}^p\) denotes text covariates, \(Y_i \in \mathbb{R}\) is the continuous outcome (average review rating), and \(T_i \in \{0, 1\}\) is the treatment assignment (presence of topic in reviews). The potential outcomes \(Y_i(0)\) and \(Y_i(1)\) represent the outcomes under control (no treatment) and treatment scenarios, respectively. The outcome \(Y_i\) is defined as \(Y_i = T_i \cdot Y_i(1) + (1 - T_i) \cdot Y_i(0)\). The goal is to estimate the Average Treatment Effect (ATE), \(\tau\), which quantifies the expected difference in outcomes due to the treatment:
\begin{equation}
\label{eq:ATE_def}
\tau = \mathbb{E}[Y_i(1) - Y_i(0)] = \mathbb{E}[Y_i(1)] - \mathbb{E}[Y_i(0)]
\end{equation}

To estimate the ATE, we consider several core assumptions and estimators: {\sl Ignorability} assumes that the treatment assignment \(T_i\) is independent of the potential outcomes, a critical condition that allows the use of a naive unbiased estimator (\(\hat{\tau}_{unadjust}\)) directly:
\begin{equation}
\label{eq:Q_unadjust}
\hat{\tau}_{unadjust} = \mathbb{E}[Y_i | T_i = 1] - \mathbb{E}[Y_i | T_i = 0]
\end{equation}
However, this assumption is often unrealistic when treatment assignment correlates with confounds. Thus, we further assume {\sl Conditional Ignorability}, which posits that the treatment assignment \(T_i\) is independent of the potential outcomes given the covariates \(X_i\). If we denote \(\mathbb{E}[Y_i | X_i = x, T_i = 1]\) as \(Q(1, x)\) and \(\mathbb{E}[Y_i | X_i = x, T_i = 0]\) as \(Q(0, x)\), then the ATE can be estimated by:
\begin{equation}
\label{eq:Q_ate}
\resizebox{0.65\columnwidth}{!}{$\hat{\tau}_{Q} = \frac{1}{n} \sum_{i=1}^{n} \left(\hat{Q}(1, X_i) - \hat{Q}(0, X_i)\right)$}
\end{equation}
Here, \(\hat{Q}(T_i, X_i)\) is the estimated response given treatment status and covariates. {\sl Positivity} assumes that every subject has non-zero probability of receiving the treatment (\(0 < P(T_i = 1 | X_i = x) < 1\) for all \(x\)). With the propensity score  \(g(x) = P(T = 1 | X = x)\), let $\hat{g}(x)$ denote the estimate of the true propensity score $g(x)$. Then the Inverse Probability Weighting (IPW) estimator is:
\begin{equation}
\resizebox{0.65\columnwidth}{!}{$\hat{\tau}_{IPW} = \frac{1}{n} \sum_{i=1}^{n} \left(\frac{T_i Y_i}{\hat{g}(X_i)} - \frac{(1 - T_i) Y_i}{1 - \hat{g}(X_i)}\right)$}
\end{equation} 
To mitigate the instability in IPW estimates due to extreme propensity scores, we also use augmented inverse propensity weighted (AIPW) estimator \cite{robins1995analysis}:
\begin{multline}
\resizebox{0.65\columnwidth}{!}{$\hat{\tau}_{AIPW} = \frac{1}{n} \sum_{i=1}^{n} \left[ \frac{T_i Y_i}{\hat{g}(X_i)} - \frac{(1 - T_i) Y_i}{1 - \hat{g}(X_i)} \right]$} \\
\resizebox{0.65\columnwidth}{!}{$- \left[ \frac{T_i - \hat{g}(X_i)}{\hat{g}(X_i)} \hat{Q}(1, X_i) - \frac{T_i - \hat{g}(X_i)}{1 - \hat{g}(X_i)} \hat{Q}(0, X_i) \right]$}
\end{multline}

\subsection{Estimating Framework}
\label{sec:estimatingframework}

We explore how specific features reflected in reviews impact the aggregate rating of the entity. To frame this as a causal inference task, we define the treatment to be any identifiable feature associated with the entity being reviewed, such as sentiments about specific aspects (e.g., `administration' for schools) or mentions of particular topics (e.g., `bullying'). For simplicity, we use a keyword-based treatment: We define a set of keywords related to the topic of interest, then categorize each entity into treatment or control groups based on the presence of these keywords in their reviews. The outcome variable is the entity's average review rating across all reviews.

\begin{figure}[t]
  \centering
  \includegraphics[width=.7\columnwidth]{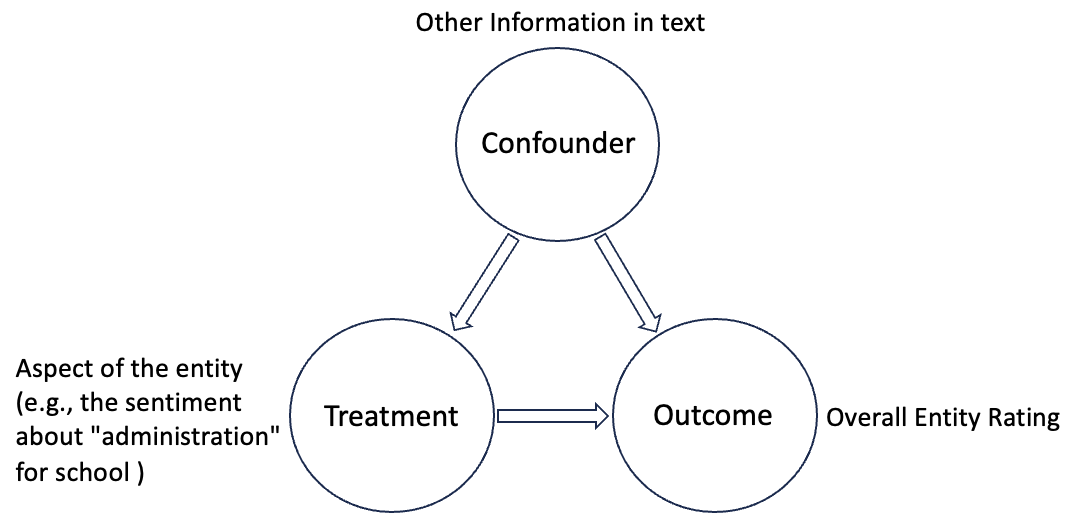}
  \caption{The causal graph of the framework}
  \label{fig:causalgraph}
\end{figure}

To address the challenges noted by \citet{pryzant2021causal} in assessing conditional ignorability -- where the treatment label may itself be influenced by other text properties -- we separate reviews that determine treatment status from those that do not. That is, all reviews that are not used to determine treatment are concatenated together to serve as covariate variables $X_i$, while the remaining reviews are discarded after being used to determine treatment.
Thus, in this causal inference task (Figure~\ref{fig:causalgraph}), average review ratings are the outcome (\(Y_i\)), and the treatment (\(T_i\)) is some real world effect determined by the presence of keywords in the reviews. We use the text from reviews without the treatment keywords as covariates (\(X_i\))—serving as a proxy for real-world factors that may influence the entity’s rating—to estimate the ATE according to Equation~\ref{eq:ATE_def}.

\subsection{CausalBERT}
\label{sec:causalbert}
A key challenge in performing causal inference with text is adjusting for confounding effects within the text. CausalBERT~\cite{veitch2020adapting}, an extension of BERT~\cite{devlin2018bert}, addresses this challenge by learning text representations that predict both the propensity score \(g(.)\) and the conditional expected outcomes \(Q(t_i, .)\), thereby learning causally sufficient text representations.

\begin{figure}[t]
  \centering
  \includegraphics[width=.7\columnwidth]{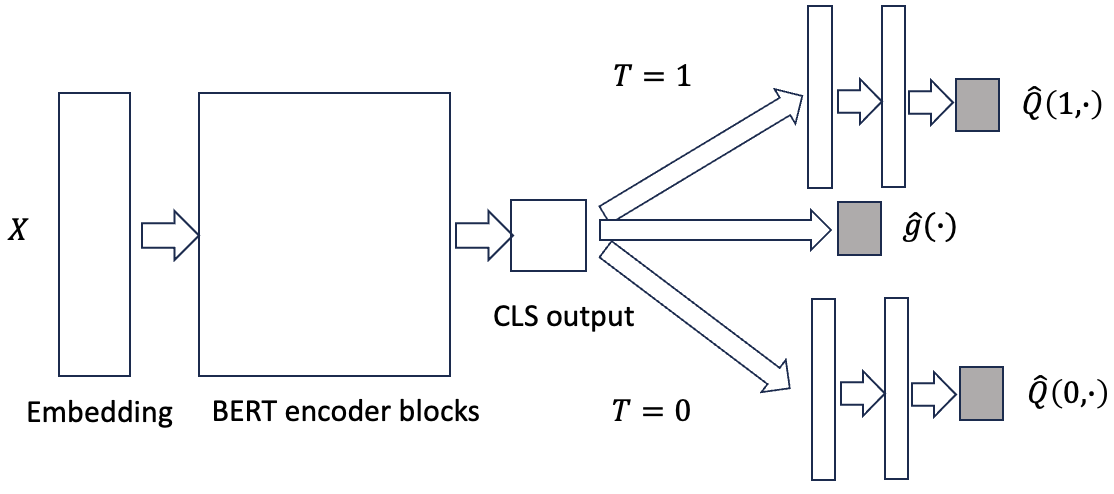}
  \caption{CausalBERT architecture.}
  \label{fig:causalbert}
\end{figure}

The architecture of CausalBERT (Figure~\ref{fig:causalbert}), inspired by Dragonnet~\cite{shi2019adapting}, processes textual data into causally relevant embeddings. Initially, CausalBERT utilizes BERT\footnote{In our project, we use pre-trained DistilBERT~\cite{sanh2019distilbert}, a smaller, faster, and lighter version of BERT base.} to transform input text \(x_i\) into a dense representation, \(h_i = \text{B}(x_i; \theta_{\text{B}})\), captured at the CLS token output. This representation is essential as it captures the critical textual information needed for causal analysis. Following the initial embedding process, CausalBERT extends into three predictive branches: (1) \(g_{nn}(h_i; \theta_g)\) for binary treatment assignment using a sigmoid-activated linear map, and (2) \(Q_{nn}^0(h_i; \theta_0)\) and \(Q_{nn}^1(h_i; \theta_1)\) for potential outcomes under non-treatment and treatment scenarios, respectively, modeled through fully connected layers (each with two hidden layers).
CausalBERT optimizes a multi-objective loss function combining mean squared error for outcome predictions and cross-entropy for treatment assignment:
\begin{align*}
\resizebox{0.99\columnwidth}{!}{$(\theta^*, \theta_{\text{B}}^*) = \arg\min_{\theta, \theta_{\text{B}}} \frac{1}{n} \sum_{i} \Big[ \left( Q_{nn}^0(h_i; \theta_0) - y_i \right)^2 (1-t_i) \nonumber$} \\
 \resizebox{0.75\columnwidth}{!}{$+ \left( Q_{nn}^1(h_i; \theta_1) - y_i \right)^2 t_i \Big] 
 + \alpha \, \text{CE}(g_{nn}(h_i; \theta_g), t_i)$}
\label{eq:causalbert_training_objective}
\end{align*}

\vspace{-.09cm}

Here, \(\theta^*\) represents the parameters of the downstream predictive models, including those for treatment assignment ($\theta_g$) and conditional expected outcome prediction ($\theta_0$, $\theta_1$). The parameter \(\theta_{\text{B}}\) corresponds to the underlying BERT parameters that are fine-tuned to optimize text representations for the specific causal inference tasks. \(\alpha\) is a hyperparameter balancing the prediction accuracies of treatment assignments and outcomes. For scalability, we do not include the `next sentence prediction' and `masked language model' tasks typically found in BERT's training regime. 

Inference involves passing data through the model to obtain propensity scores $\hat{g}(.)$ and potential outcomes $\hat{Q}(0, .)$, $\hat{Q}(1, .)$, which are then used in the ATE estimators from Section~\ref{subsec:estimation-treatment-effects}. We next describe several enhancements to refine the performance of CausalBERT.

\subsection{Temperature Scaling in CausalBERT}
\label{sec:temperature-scaling}
In observational studies,  treatment and control groups often exhibit a lack of complete overlap in confounder distributions, which hinders the derivation of empirical counterfactuals~\cite{gelman2021regression}. In CausalBERT, this lack of overlap can arise due to poorly calibrated propensity scores, where extreme $\hat{g}(.)$ values make adjustments unstable. 
To address this issue, we enhance CausalBERT with Temperature scaling \cite{guo2017calibration}, which introduces a temperature parameter \(M > 0\) to adjust the confidence levels of the propensity score predictions. This parameter helps align the predicted probabilities with their actual confidence levels, mitigating the risk of extreme propensity scores that could negatively affect subsequent estimations such as IPW or AIPW.

In CausalBERT, for a given logit vector \(z\) from the treatment  prediction branch, the adjusted confidence prediction is: $
    \hat{g}_{scaled} = \max_k \sigma_{\text{SM}}\left(\frac{z_k}{M}\right)$,
where \(\sigma_{\text{SM}}\) denotes the softmax function and \( k \) denotes the class index. The temperature \(M\) (where \(M > 1\)) ``softens'' the  probabilities, increasing output entropy and leading to more uniform class probabilities. Conversely, as \(M\) approaches zero, the softmax probabilities converge to a point mass, favoring more confident predictions.

To determine \( M \), we minimize the Negative Log-Likelihood (NLL) of the propensity score predictions on a heldout validation set: $
    M^* = \arg\min_{M} \text{NLL}\left(\frac{z}{M}; t\right)
$,
where \( t \) is the true treatment label. Adjusting \( M \) does not change the predicted class — it only refines the softmax probabilities to better represent the underlying uncertainties in treatment assignment predictions.


\subsection{Mitigating Overadjustment}
\label{subsec:mitigating-overadjustment}
Overadjustment for potential confounds can lead to biased effect estimates
~\cite{vanderweele2009relative}. In the CausalBERT objective, $\alpha$ determines the importance placed on the treatment prediction head, which in turn can influence the amount of confounder adjustment. We propose setting $\alpha$ based on estimates of the amount of confounding in the data.
%
To estimate the amount of confounding, we use the accuracy of treatment prediction as a signal.
This assumes that lower treatment classification accuracies generally indicate weaker confounding. In such cases, we can increase \( \alpha \), thereby intensifying the model’s focus on treatment classification without the risk of substantial bias from confounds. 
By correlating \( \alpha \)  with observed treatment accuracy, we employ an empirical approach to adjust $\alpha$, enhancing causal effect estimation across confounding scenarios. We explore this further in \S\ref{sec:results}.

\subsection{Interpreting CausalBERT}
\label{sec:interpretingcausalbert}
As true causal effects are rarely known, it is important to have qualitative methods to assess the validity of CausalBERT. We explore two qualitative methods to do so: \textbf{CLS Comparative Analysis} and \textbf{Integrated Gradients}.

First, building on interpretability methods for deep learning, our \textbf{CLS Comparative Analysis} quantifies the aggregate attention for the CLS token and compares the fine-tuned CausalBERT with baseline DistilBERT to determine how fine-tuning affects term importance. We analyze tokens that significantly influence the CLS token -- excluding stopwords and punctuation -- using two strategies. \textbf{General Top Contributing Tokens} ranks tokens by attention score, selecting the most influential for each document, while \textbf{Max Subarray for Continuous Contribution} identifies contiguous token subarrays with maximum influence on the CLS token.
By aggregating these tokens across all documents, we compare the top influential tokens between CausalBERT (\(A\)), which controls for confounding effects through its design, and DistilBERT (\(B\)), which does not, using \(A \setminus B\) to assess changes in attention due to fine-tuning. Additionally, in semi-synthetic experiments below, we assess the proportion of \(A \setminus B\) tokens that correspond to the confounding variable we inject into the data, providing an additional check that CausalBERT is discovering confounders appropriately.

Second, we employ \textbf{Integrated Gradients (IG)}~\cite{sundararajan2017axiomatic}, an interpretability technique that attributes the prediction of a deep learning model to its input features.
For each output component of CausalBERT (treatment and outcomes), IG identifies the tokens that significantly increase or decrease the model’s predictions. By aggregating across instances, we compile the most influential tokens for each prediction task. We denote $g^+$ and $g^-$ as the top terms that respectively increase and decrease the propensity score prediction, while $Q_{0}^+$ and $Q_{1}^+$ describe the top terms that enhance the outcome predictions for the control and treated groups, and $Q_{0}^-$ and $Q_{1}^-$ for those that diminish these predictions. Each token in these categories is associated with a contribution weight that quantifies its impact on the model's output. These weights are normalized within each respective list to highlight the relative importance of each term.

\section{Experiments}
\label{sec:experiments}
We empirically examine the capabilities of CausalBERT and the proposed enhancements, using both real and semi-synthetic review data, with a focus on the following questions.
\textbf{RQ1:} How does CausalBERT performance vary with confounder strength? \textbf{RQ2:} What effect does Temperature Scaling have on treatment effect estimation? \textbf{RQ3:} How does hyperparameter \( \alpha \) in the loss function influence  overadjustment and how does its optimal value relate to treatment prediction accuracy? \textbf{RQ4:} How effective are interpretability methods at surfacing the confounders discovered by CausalBERT? \textbf{RQ5:} To what extent do educational aspects, such as `bullying' and `administration,' impact overall school ratings in real-world data? 

\vspace{-.1cm}

\paragraph{Dataset}
\label{sec:Dataset}
We analyze 677,210 reviews from GreatSchools.org \footnote{The dataset was provided by our collaborator GreatSchools and is not publicly available.} , covering 83,795 public, private, and charter schools in the United States between 2002-2019. 
We investigate the impact of school-related topics such as `bullying,' `academic performance,' `administration,' `extracurricular activities,' and `curriculum,' each defined by a keyword list established by prior work~\cite{harris2022,gillani2021parents} (Appendix~\ref{sec:keywords_list}). For each topic, we first separate reviews into those that discuss the topic and those that do not.
For outcome, we normalize review ratings by computing state-specific z-scores and average these scores for each school over the selected period. Thus, outcome values are in standard units. For treatment assignment, we adopt different methods based on the nature of the topics discussed in the reviews. For neutral topics such as `administration,' treatment is determined by sentiment about that topic. In this case, an entity is considered treated ($T=1$) if all relevant reviews express positive sentiment and untreated ($T=0$) if all are negative. Schools with a mix of positive and negative reviews on administration are excluded from the analysis to maintain a clear treatment distinction. For the negative topic of `bullying,' treatment is  1 if the topic is mentioned, 0 otherwise.

In our framework, the `bullying' task aims to assess how the presence of bullying affects the school's overall ratings. For other topics like `administration,' we treat the sentiment of text referencing administration as a proxy for real-world administrative quality and measure how that factor's sentiment influences the school's overall rating. Our causal pathway assumes multiple school attributes can affect overall ratings, so we isolate the effect of a specific aspect by controlling for other factors. For instance, to examine bullying, we separate it from other negative conditions—like poor administration—that might also lower ratings. 
We have between 3,900 and 13,300 schools for each topic; more detailed statistics (e.g., average total reviews per school) are provided in \S\ref{app:school_stats}.
 
\paragraph{Semi-Synthetic Data Setup}
\label{sec:Semi-synthetic data_setup}

First, we employ a semi-synthetic evaluation framework \cite{weld2022adjusting} to evaluate CausalBERT's treatment effect estimation capabilities using the bullying topic. We simulate a binary confound $C_i \in \{1, 2\}$ by inserting text related to an academic challenges topic into certain reviews (see \S\ref{subsec:academic-challenge-templates} for terms). Schools in Class 1 receive these injected sentences, while those in Class 2 do not.

To manipulate the ATE, we vary the true ATE by defining two outcome models based on treatment status: for Class 1, outcomes are modeled as \(Y \sim \mathcal{N}(u_2, 0.3)\) when treatment \(T=1\) and \(Y \sim \mathcal{N}(u_1, 0.3)\) when \(T=0\). Class 2 maintains uniform effects with \(Y \sim \mathcal{N}(u_2, 0.3)\), indicating no treatment effect from textual confounding. Here, \(u_2\) is set at -0.3, and \(u_1\) varies within $\{0.3, 0.4, 0.5\}$, creating corresponding true ATE \(u\) values of $\{-0.3, -0.35, -0.4\}$.

The confounder strength is controlled by adjusting the probability \(p\), which defines the treatment assignment probabilities within each class. We vary \(p\) from 0.9 to 0.5, where for Class 1: \(P(T=1 | C=1) = 1-p\) and \(P(T=0 | C=1) = p\), and inversely for Class 2: \(P(T=1 | C=2) = p\) and \(P(T=0 | C=2) = 1-p\).

In our semi-synthetic experiments, we construct data according to specific values for $u$ and $p$ by sampling schools to satisfy these constraints. Thus, the only synthetic part of these experiments is the sampling procedure and the injection of confounding sentences. Each experiment samples 5,000 instances from a population of 13,361.

\paragraph{Evaluation Metrics}
\label{sec:Evaluation_Metrics}
We perform 5-fold cross-validation for the semi-synthetic data experiments and bootstrap aggregation for real-world data analysis (\S\ref{sec:data_collection_detail}). Estimators $\hat{\tau}_{Q}$, \(\hat{\tau}_{IPW}\), and \(\hat{\tau}_{AIPW}\), and their calibrated versions, are used to estimate the ATE, with the naive estimator \(\hat\tau_{unadjust}\) as a baseline. Since we have different true causal treatment effect designs, we use the error ratio to evaluate the estimation results, defined as
$
|\hat{\tau}_{\text{est}} - \tau_{\text{true}}| / \tau_{\text{true}}
$,
where \(\hat{\tau}_{\text{est}}\) is the estimated treatment effect and \(\tau_{\text{true}}\) is the true treatment effect. We also report the accuracy of treatment prediction and the mean squared error (MSE) of outcome prediction.

\section{Results}
\label{sec:results}

\subsection{Evaluation on Semi-Synthetic Data}

\paragraph{Treatment Assignment Prediction} Table~\ref{tab:accuracy_summary} reports average treatment prediction accuracy (\(\alpha\)=0.33) for different confounder strengths $p$ and true ATE $u$. For every fixed $u$, the accuracy of treatment assignment prediction increases linearly as the confounder strength increases. These results provide evidence that CausalBERT can capture the relationship between treatment and confounder variables. 
Furthermore, the results suggest that treatment prediction accuracy can serve as a reliable indicator of confounder strength in real-world data, which we will return to below.

\begin{table}[ht]
\centering
\resizebox{1.01\columnwidth}{!}{
\setlength{\tabcolsep}{3pt} 
\begin{tabular}{@{}cccc|ccccc@{}}
\toprule
& \multicolumn{3}{c|}{Treatment Accuracy}  & \multicolumn{3}{c}{MSE}\\ 
\textbf{$p$} & $u=\text{-}.40$ &$u=\text{-}.35$ & $u=\text{-}.30$ & $u=\text{-}.40$ & $u=\text{-}.35$ & $u=\text{-}.30$ \\ 
\midrule
.5 & .59$\pm$.01 & .58$\pm$.03 & .57$\pm$.02 & .077$\pm$.01 & .069$\pm$.00 & .080$\pm$.01 \\
.6 & .62$\pm$.02 & .62$\pm$.01 & .61$\pm$.02 & .088$\pm$.01 & .074$\pm$.00 & .085$\pm$.01 \\
.7 & .67$\pm$.02 & .67$\pm$.02 & .63$\pm$.01 & .067$\pm$.00 & .077$\pm$.01 & .085$\pm$.02 \\
.8 & .77$\pm$.01 & .74$\pm$.01 & .73$\pm$.02 & .066$\pm$.01 & .071$\pm$.01 & .070$\pm$.00 \\
.9 & .83$\pm$.02 & .82$\pm$.02 & .83$\pm$.01 & .068$\pm$.01 & .069$\pm$.01 & .075$\pm$.01 \\
\bottomrule
\end{tabular}
}
\caption{Average treatment classification accuracy and outcome MSE using CausalBERT(\(\alpha\)=0.33) for different confounder strength $p$ and true ATE $u$.}
\label{tab:accuracy_summary}
\end{table}

\paragraph{Outcome Prediction} 

Table~\ref{tab:accuracy_summary} also shows the average MSE of outcome prediction. Unlike treatment accuracy, MSE does not exhibit sensitivity to confounder strength. This suggests that the regression task of predicting outcomes is less challenging for CausalBERT compared to the classification task of treatment prediction, indicating different levels of complexity and sensitivity in these tasks.

\paragraph{Effects of Confounder Strength}

\begin{figure}[htp] 
\centering
\includegraphics[width=.85\linewidth]{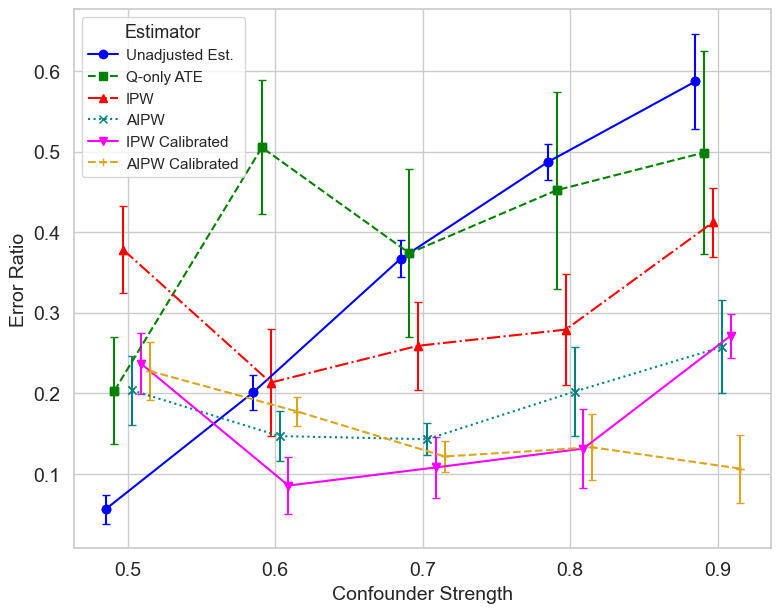}
\caption{Average error ratio (with standard error) of treatment effect estimation by confounder strength.} 
\label{fig:ate_est}
\end{figure}

Figure~\ref{fig:ate_est} shows the performance of different estimators across various confounder strengths for semi-synthetic datasets with a fixed true ATE of -0.3 (see Appendix for similar results with true ATE values -0.35 and -0.4). The baseline unadjusted ATE estimator performs better as confounder strength decreases, achieving optimal results at a confounder strength of $0.5$. This observation aligns with expectations for a randomized trial scenario, where treatment assignment is completely random and devoid of confounding biases, thus rendering this naive estimator unbiased.

In contrast, the Q-only, IPW, and AIPW estimators significantly outperform the baseline when confounder strength is high ($p > 0.7$).  Both IPW and AIPW outperform Q-only in scenarios with strong confounders, suggesting that the propensity scores calculated by CausalBERT are particularly beneficial in aiding robust ATE estimation. 

Calibrated IPW and AIPW estimators display the most consistent performance across various settings, suggesting that temperature scaling, which aligns predicted probabilities more closely with their true confidence levels, enhances reliability. However, under conditions of weak confounder strength ($p = 0.5$), CausalBERT tends to underperform compared to the unadjusted estimator. This could be due to the model capturing irrelevant information perceived as confounding. Next we will further explore these observations on temperature scaling and overadjustment.

\paragraph{Effect of Temperature Scaling}

\begin{figure*}
\begin{minipage}[t]{.34\textwidth}
  \centering
  \includegraphics[width=.99\linewidth]{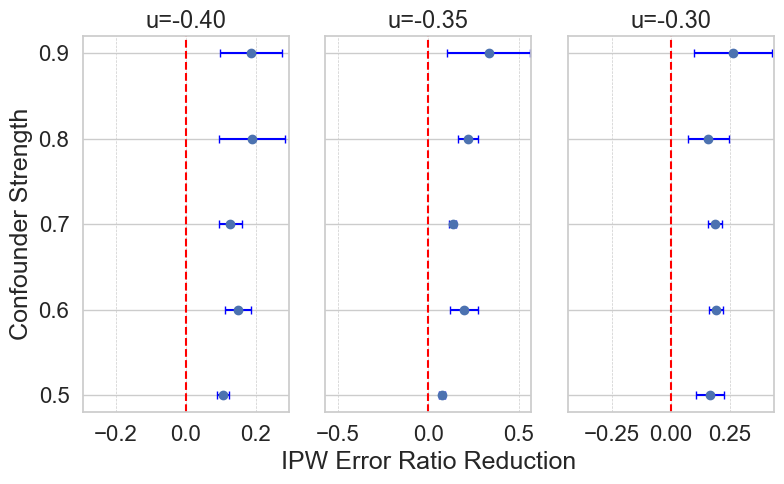}
\end{minipage}%
\hfill
\begin{minipage}[t]{.35\textwidth}
  \centering
  \includegraphics[width=.9\linewidth]{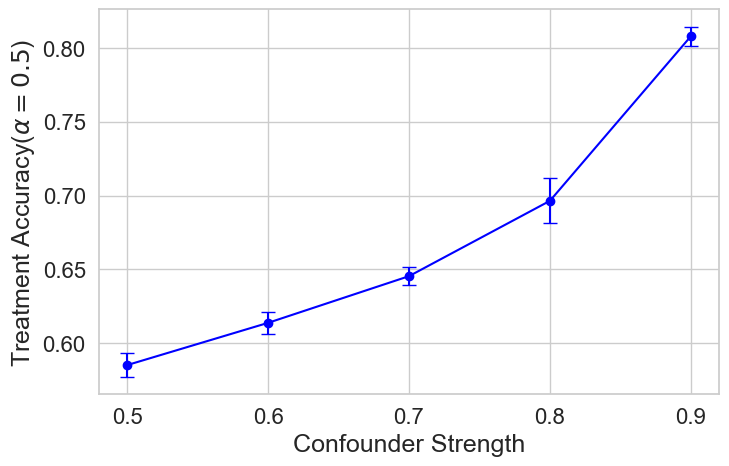}
\end{minipage}%
\hfill
\begin{minipage}[t]{.30\textwidth}
  \centering
  \includegraphics[width=.99\linewidth]{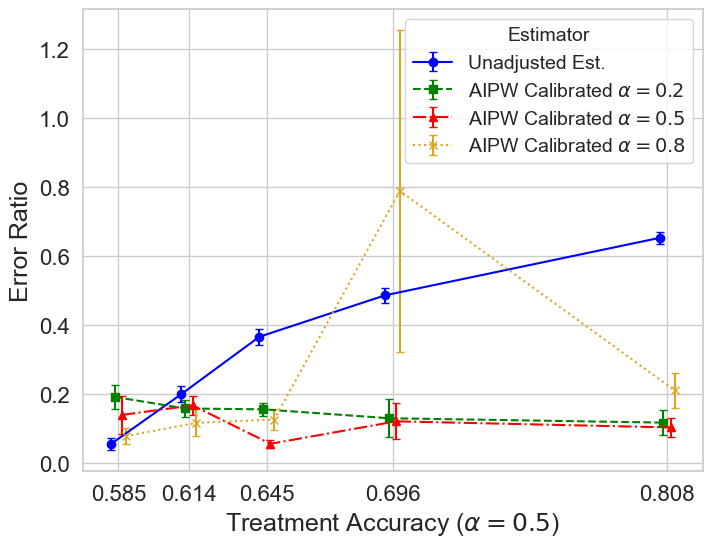}
\end{minipage}
\caption{(Left) Error ratio decrease by temperature scaling on IPW. (Center) Treatment accuracy (standard error) for CausalBERT (\(\alpha\)=0.5) by confounder strength. (Right) Error ratio by treatment accuracy and $\alpha$  (trained with $\alpha = 0.5$) on a semi-synthetic dataset with a fixed true ATE $u=-0.3$, for the AIPW Calibrated model.}
\label{fig:combined}
\end{figure*}

Figure~\ref{fig:combined} (Left) presents the average error reduction provided by temperature scaling on estimates from IPW ATE (\(\alpha=0.4\)) across various confounder strengths and true ATEs. Temperature scaling consistently enhances IPW estimates by aligning predicted probabilities more closely with their actual confidence levels,  mitigating the issue of extreme propensity scores. This is evident as scaled IPW consistently shows reduced error ratios compared to non-scaled IPW across all configurations. For AIPW (see Appendix), the benefit of scaling is less pronounced and primarily observed in high confounder strength scenarios. AIPW's inherent double robustness mechanism may make it less susceptible to the pitfalls of extreme propensity scores. 

Both IPW and AIPW show significant improvement from scaling at a confounder strength of 0.9. This may be attributed to the propensity model's increased risk of producing extremely confident predictions (too close to 0 or 1) at high treatment probabilities, where the calibration can have a substantial impact. For a further analysis, see \S\ref{sec:propensity_score_balance}.

\paragraph{Mitigating overadjustment} 

Figure~\ref{fig:combined} (Center) shows the linear relationship between confounder strength and treatment prediction accuracy, trained with \( \alpha = 0.5 \). This analysis suggests that treatment prediction accuracy can serve as an indicator of the amount of confounding within a dataset. Building upon this relationship, Figure~\ref{fig:combined} (Right) explores the impact of varying $\alpha  \in \{0.2, 0.5, 0.8\}$ on model performance for calibrated AIPW on a fixed true ATE of \( u = -0.3 \). (See Appendix for IPW results.) Taken together, these results suggest that we can guide our selection of $\alpha$ based on treatment prediction accuracy. Notably, at lower treatment accuracies, indicative of weaker confounding, \( \alpha = 0.8 \) yields the best ATE estimates. Conversely, in regions of higher treatment accuracy, \( \alpha = 0.5 \) performs best. This analysis can guide the selection of an appropriate \( \alpha \) value: a higher \( \alpha \) when treatment accuracy is around or below 0.6 and a moderate \( \alpha \) when treatment accuracy exceeds 0.65.

A possible explanation for the effectiveness of a higher \(\alpha\) in low-confounding scenarios is that it shifts the model’s focus towards treatment classification, enhancing its sensitivity to treatment signals. In these settings, clearer signals emerge because the primary challenge is not adjusting for confounds but accurately identifying treatment presence. Thus, by prioritizing treatment prediction, the model more effectively captures and learns from these direct treatment effects, improving its performance in estimating treatment impacts.


\paragraph{CLS Comparative Analysis} 
We now turn to a qualitative analysis to investigate which words and phrases drive the predictive signal in CausalBERT. The analyses in this section are conducted across semi-synthetic datasets, fixing the true ATE \(u=-0.3\), setting $\alpha=0.4$, and varying confounder strengths from 0.9 to 0.5.

To understand which factors are potential confounds discovered by CausalBERT, Table~\ref{tab:cls_word_focus}, displays results from applying the CLS comparison method while varying confounder strength. We observe that when the confounder strength is strong  ($p=.9$ or $.8$), the tokens in \(A \setminus B\) predominantly derive from our inserted confounder text, indicating that supervised fine-tuning enhances the model’s emphasis on confounder information within the text representation. When the confounder strength is weak, however, the model shifts its focus towards the treatment itself, with tokens such as `bull,' `horrible,' `rude,' and `bad' emerging prominently. 
\begin{table}[!ht]
\centering
\small
\resizebox{0.49\textwidth}{!}{
\begin{tabularx}{.55\textwidth}{cXc}
\toprule
\textbf{$p$} & \multicolumn{1}{c}{\textbf{Top terms}} & \textbf{Prop.} \\
\midrule
$.9$ & \textbf{\#\#pf}, \textbf{exams}, sweet, special, involvement, involved, \textbf{handling}, \textbf{\#\#istic}, \textbf{course}, make, \textbf{structured}, \textbf{study}, want, kind, \textbf{thinking}, \textbf{transferring}, \textbf{\#\#hel}, \textbf{sy}, even, \textbf{face}, \textbf{understand}, \textbf{challenges}, \textbf{\#\#ul}, \textbf{\#\#bus}, \textbf{believe}, know, \textbf{semester}, environment, \textbf{advice}, \textbf{professor}
 & .667 \\
\hline
$.8$ & simply, \textbf{\#\#pf}, \textbf{exams}, deeply, \textbf{\#\#real}, \textbf{handling}, \textbf{course}, \textbf{deal}, \textbf{sizes}, \textbf{materials}, level, \textbf{feedback}, caring, \textbf{transferring}, \textbf{assignments}, focused, elementary, oldest, \#\#ted, \textbf{sy}, gifted, \textbf{face}, \textbf{understand}, \textbf{poorly}, \textbf{challenges}, \textbf{assistants}, \textbf{\#\#ul}, awesome, growing, job, \textbf{large}, \textbf{believe}, \textbf{semester}, amazing, \textbf{advice}, smile, \textbf{professor}
 & .622 \\
\hline
$.5$ & told, needs, \#\#ing, experience, middle, last, people, bad, need, never, \#\#t, \textbf{un}, nothing, even, horrible, rude, go, bull, \#\#ied
 & .053 \\
\bottomrule
\end{tabularx}
}
\caption{Terms with high attention in CausalBERT but not DistilBERT by confounder strengths $p$. Bold terms were inserted by the synthetic confounder. Prop. is fraction of top tokens from the confounder template.}
\label{tab:cls_word_focus}
\end{table}

These tokens are closely associated with the manifestation of bullying, suggesting that CausalBERT is sensitive to direct treatment signals when the confounder-treatment correlation is minimal. These results indicate that CausalBERT effectively identifies confounding topics when the confounding strength is high, but has more difficulty doing so when confounding strength is low.

\subsection{Application to Original Data}
\label{sec:real-data-application}
\begin{figure}[t]
  \includegraphics[width=\columnwidth]{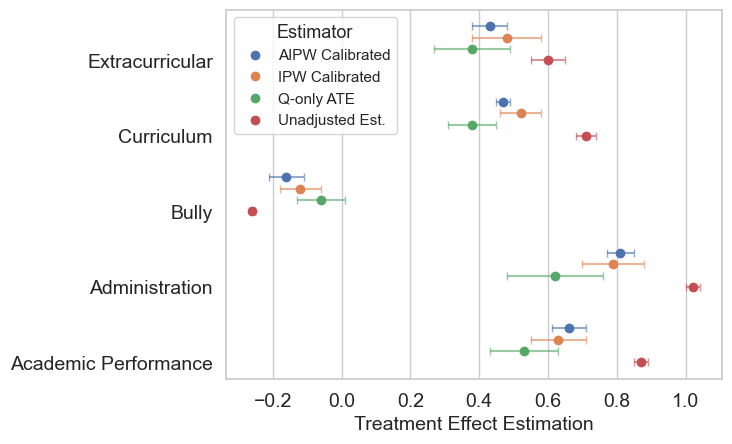}
  \caption{ATEs by topic (\(\alpha\)=0.5).}
  \label{fig:real_est}
\end{figure}
\begin{figure}[t]
  \includegraphics[width=\columnwidth]{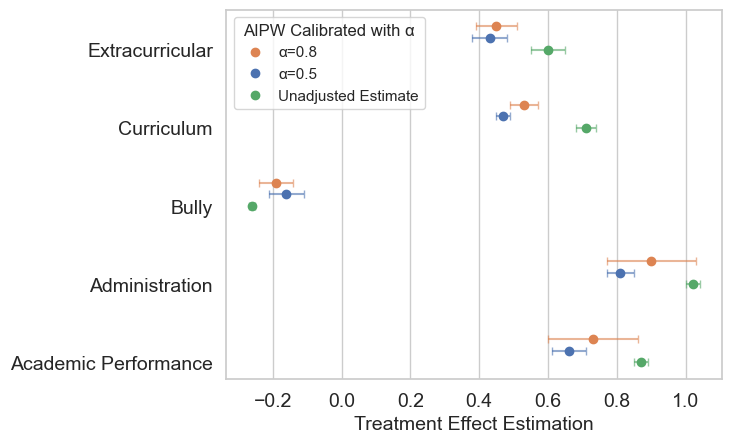}
  \caption{AIPW Calibrated estimates by topic and \(\alpha\) .}
  \label{fig:real_est_alpha}
\end{figure}

 We next apply CausalBERT to the original school review data, without any synthetic data injection. Figure~\ref{fig:real_est} presents the bootstrapped treatment effect estimates by topic with $\alpha=0.5$. We observe that our adjustment for confounders consistently reduces the magnitude of the effect estimates when compared to the unadjusted estimates. 
 For instance, in the case of `bullying', AIPW-Calibrated produces an effect of -0.16, in contrast to a more substantial decrease of -0.26 in the unadjusted estimate. This difference suggests the presence of confounding factors, such as poor administration, that often co-occur with bullying incidents, lead to a compounded negative impact on school ratings. Similarly, in `academic performance', where AIPW-Calibrated shows a treatment effect of +0.66 compared to the unadjusted effect of +0.87, positive academic performance may coexist with other favorable conditions, such as educational programs and community involvement. 

Comparing the overall effect sizes,  `administration' has the highest effect, followed by `academic performance,' `curriculum`, and `extracurricular activities.'  In contrast, 'bullying' has a negative impact, as expected. We note that it may be difficult to directly compare the magnitudes of the bullying topic with the others, due to the difference in how the treatment categories were determined (\S\ref{sec:Dataset}).

\subsubsection{Integrated Gradients}
\begin{table}[!ht]
\centering
\small
\resizebox{0.49\textwidth}{!}{
\begin{tabularx}{.6\textwidth}{XX}
\toprule
\multicolumn{2}{c}{\textbf{$g^+$}~~~~~~~~~~~~~~~~~~~\textbf{Bullying}~~~~~~~~~~~~~~~~~~~\textbf{$g^-$}}\\
\midrule
horrible, terrible, rude, worst, \#\#ing, un, bad, lack, bull, seems, office, needs, nothing, \#\#t, negative, different, disappointed, many, \#\#ied, problem &   great, love, best, wonderful, amazing, excellent, loves, every, caring, awesome, much, top, say, happy, truly, family, dedicated, community, support, pleased \\
\midrule
\multicolumn{2}{c}{\textbf{$g^+$}~~~~~~\textbf{Academic Performance}~~~~~~\textbf{$g^-$}}\\
\midrule
great, love, school, excellent, amazing, best, program, staff, happy, community, know, always, wonderful, make, learn, really, highly, part, awesome, dedicated& principal, get, administration, needs, horrible, education, new, want, rude, worst, care, leadership, bad, nothing, terrible, bullying, feel, reviews, di, bull \\
\bottomrule
\end{tabularx}
}
\caption{Top 20 terms for treatment assignment prediction ($g^+$ and $g^-$) by Integrated Gradients.}
\label{tab:integrated_gradients_real}
\end{table}

Table~\ref{tab:integrated_gradients_real} presents the top 20 terms for $g^+$ and $g^-$ by applying Integrated Gradients for `bullying' and
`academic performance' topics, highlighting how specific terms influence treatment prediction. For `bullying,' terms like `horrible', `terrible', and `rude' significantly increased the propensity score, indicating that bullying often co-occurs in schools with many other negative reviews. Conversely, positive terms such as `great', `happy', `love', `support', `family', and `community' decreased the propensity score, reflecting environments where bullying is likely absent and the school
atmosphere is perceived positively. For `academic performance', positive sentiment terms are associated with schools perceived favorably academically, while administrative terms like `principal' and `administration' correlate with negative perceptions of academic performance. This pattern indicates that perceptions of academic quality are correlated not only with leadership and administrative factors but also with community engagement and the quality of educational programs.

\vspace{-.1cm}
\paragraph{Selecting $\alpha$}
To apply CausalBERT to real data, we assess the confounder strength by first applying CausalBERT with a moderate \(\alpha\) to estimate treatment prediction accuracy. The relationship between this initial accuracy and the estimation performances, as in Figures~\ref{fig:combined} (Center), informs the selection of an appropriate \(\alpha\) for subsequent analysis. Once an optimal \(\alpha\) is determined, we recommend re-running CausalBERT with this adjusted \(\alpha\) and employing temperature scaling to enhance the robustness of the estimate.

Applying this approach, we find that treatment prediction accuracies by topic are: bullying=.53, extracurricular=.59, curriculum=.62, academic performance=.63, administration=.65, suggesting low to moderate confounder strength across all topics. According to Figure~\ref{fig:combined} (Right), we repeat the experiment with a higher \(\alpha=0.8\), using AIPW Calibrated estimator, since it is the most robust in our semi-synthetic experiments. The results in Figure~\ref{fig:real_est_alpha} show that the estimated treatment effects are consistently larger than those at \(\alpha=0.5\) and closer to the unadjusted estimates. This new result with \(\alpha=0.8\) may offer more accurate estimates, as prior experiments with semi-synthetic data showed that, under weak confounder conditions,  $\alpha=0.8$ tended to be the most precise.

\section{Related Work}
\label{sec:related_work}

\vspace{-.1cm}

Aspect-based sentiment analysis analyzes sentiment toward specific topics~\cite{hu2004mining,tang-etal-2015-document,ruder-etal-2016-hierarchical},
yet generally does not address the causal impact of these aspects on overall ratings, potentially conflating correlation with causation. 
There is growing research in causal inference with text~\cite{keith2020text,feder2021causalm,keith2023rct,veljanovski2024doublelingo}. 
A common approach is to use NLP to identify confounds and adjust for them in estimation~\cite{sridhar2019estimating,roberts2020adjusting,mozer2020matching}. Other methods method adjust for confounders from text with supervised NLP. \citet{veitch2020adapting} introduced CausalBERT, leveraging pre-trained BERT models~\cite{devlin2018bert} to derive ``sufficient" embeddings that capture confounding properties within texts, optimizing predictions for both treatment and counterfactual outcomes. While they highlighted the potential for deep learning methods to improve estimation accuracy, they also outlined several future challenges: refining deep learning approaches to enhance estimation accuracy; developing visualization and sensitivity analysis tools to clarify the ``black box'' nature of embeddings; and expanding semi-synthetic simulations into a comprehensive benchmarking strategy. These challenges have inspired the work in this paper.

\vspace{-.1cm}

\section{Conclusions}
\label{sec:conclusions}
In this study, we have extended CausalBERT to understand how factors mentioned in school reviews affect overall ratings. Through semi-synthetic experiments, we verified the effectiveness of CausalBERT in a controlled setting, which then guided our application to real-world data. Our analysis indicates that Temperature Scaling and Integrated Gradients can refine causal estimates and enhance interpretability. 
Analysis of U.S. K-12 school reviews found that educational aspects like `Administration' and `Academic Performance' have significant influence on school ratings.

\section{Limitations}
\label{sec:limitations}
Like most studies involving causal inference, true effects are unknown, and thus there is unavoidable uncertainty in the results.  
Despite rigorous validation using semi-synthetic data that demonstrates the model's effectiveness in controlled scenarios, the extrapolation of these results to real-world data must be treated with caution. 

Additionally, our reliance on keyword-based treatment identification introduces another layer of potential noise. This method assumes that the presence of predefined keywords, such as 'bullying', sufficiently identifies relevant reviews. 

\section{Ethics Statement}
\label{sec:ethics_statement}
Our analysis focuses on publicly available online school reviews. While we are primarily interested in understanding specific school-related topics—such as "bullying," "academic performance," "administration," "extracurricular activities," and "curriculum" that influence overall school ratings, our work could also be utilized by administrators or parents, potentially leading to unintended consequences for certain schools. We caution against over-reliance on our results and emphasize considering each school's unique context. All data were provided by our collaborator, GreatSchools, in an anonymized form, containing no personally identifiable information. We only use aggregated, school-level data; any individual-level identifiers, such as reviewer names or addresses, were removed prior to our access.

\section*{Acknowledgments}
This work was supported in part by the Harold L. and Heather E. Jurist Center of Excellence for Artificial Intelligence at Tulane University and the Tulane University Center for Community-Engaged Artificial Intelligence. Linsen Li was supported by NSF Award IIS-III-2107505. Aron Culotta was supported by NSF Awards IIS-HCI-2333537 and SCC-IRG-2427237.  Nicholas Mattei was supported by NSF Awards IIS-RI-2007955, IIS-III-2107505, IIS-RI-2134857, IIS-RI-2339880 and CNS-SCC-2427237. 

We would like to thank Douglas N. Harris and Jamie M. Carroll in their help with this work. Portions of this research were carried out under the auspices of the National Center for Research on Education Access and Choice (REACH) based at Tulane University, which is supported by the Institute of Education Sciences, U.S. Department of Education, through Grant R305C100025 to The Administrators of the Tulane Educational Fund. The opinions expressed are those of the authors and do not represent views of the Institute, the U.S. Department of Education, Great Schools, or any other organization.

\clearpage
\bibliography{causal-text}

\clearpage
\appendix
\section{Technical Appendix}
\label{sec:appendix}

\subsection{Schools Per Topic Table}\label{app:school_stats}

\begin{table}[ht]
\centering
\resizebox{0.7\columnwidth}{!}{
\setlength{\tabcolsep}{5pt} 
\small 
\begin{tabular}{@{}rccc@{}}
\toprule
\textbf{topic}       & $T=1$ & $T=0$ & \textbf{total}\\
\midrule
bullying             & 4,688 & 8,673 & 13,361\\
administration       & 2,940 & 1,049 & 3,989\\
academic performance & 4,111 & 1,670 & 5,781\\
extracurricular      & 5,406 & 1,385 & 6,791\\
curriculum           & 4,431 & 2,063 & 6,494\\
\bottomrule
\end{tabular}
}
\caption{Schools in treatment and control groups by topic.}
\label{tab:school_stats}
\end{table}

\begin{table}[ht]
\centering
\resizebox{\columnwidth}{!}{
\setlength{\tabcolsep}{5pt}
\small
\begin{tabular}{lcccc}
\toprule
\textbf{Topic} & \multicolumn{2}{c}{\textbf{Total Reviews}} & \multicolumn{2}{c}{\textbf{Keyword-Containing Reviews}} \\
 & $T=1$ & $T=0$ & $T=1$ & $T=0$ \\
\midrule
\textbf{Bullying} & 11.08 $\pm$ 9.06 & 8.34 $\pm$ 5.23 & 1.60 $\pm$ 1.10 & 0.00 $\pm$ 0.00 \\
\textbf{Administration} & 8.09 $\pm$ 5.34 & 6.62 $\pm$ 2.14 & 3.41 $\pm$ 2.87 & 2.35 $\pm$ 1.39 \\
\textbf{Academic Performance} & 8.01 $\pm$ 4.76 & 6.79 $\pm$ 2.37 & 2.39 $\pm$ 1.74 & 1.77 $\pm$ 0.98 \\
\textbf{Extracurricular} & 9.64 $\pm$ 6.55 & 8.22 $\pm$ 4.12 & 1.82 $\pm$ 1.35 & 1.27 $\pm$ 0.58 \\
\textbf{Curriculum} & 8.61 $\pm$ 5.74 & 7.37 $\pm$ 2.90 & 2.05 $\pm$ 1.69 & 1.48 $\pm$ 0.80 \\
\bottomrule
\end{tabular}
}
\caption{Average ($\pm$ standard deviation) of total and keyword-containing reviews for treated ($T=1$) and untreated ($T=0$) per schools by topic.}
\label{tab:review_stats}
\end{table}

\subsection{Academic Challenge Post Templates}
\label{subsec:academic-challenge-templates}
This subsection outlines the templates and words used to generate academic challenge posts for our semi-synthetic data generation process. We use the following sentence templates to simulate academic challenges faced by students, with placeholders indicated by \{\}:
\begin{itemize}
    \item "I can't believe I have to deal with \{\} in this course."
    \item "Every semester, I face \{\} in my classes."
    \item "The professor doesn't understand the challenges of \{\}."
    \item "Does anyone have advice on handling \{\} in school?"
    \item "I'm thinking of transferring because of \{\}."
\end{itemize}

The placeholders are filled with words representing academic challenges: "unrealistic assignments", "difficult exams", "lack of study materials", "unhelpful teaching assistants", "large class sizes", "lack of feedback", "poorly structured syllabus".

\textbf{Data Generation Procedure:}
Each synthetic data point is generated by randomly selecting one of the above words and inserting it into a randomly chosen template. This process introduces variability and simulates real-world academic challenges, facilitating the evaluation of CausalBERT's performance in handling text-based confounders.

\subsection{Estimator Comparison}
\label{subsec:estimationtreatmenteffectsappendix}
In this section, we provide additional insights into the comparison of various treatment effect estimators used in our analysis. 

\textbf{Naive Estimator (\(\hat{\tau}_{unadjust}\)):} The simplest approach, assumes ignorability, where the treatment is independent of potential outcomes. While straightforward, this estimator is often unreliable in practice due to potential confounding variables that correlate with treatment assignment.

\textbf{Q-only Estimator (\(\hat{\tau}_{Q}\)):} It refines the Naive Estimator by assuming conditional ignorability, where treatment assignment is independent of potential outcomes given covariates \(X_i\). This method adjusts for confounders but depends heavily on accurately modeling the relationship between covariates and outcomes, which can be challenging with varying covariate distributions across treatment and control groups\cite{rubin2001using}.

\textbf{Inverse Probability Weighting (IPW):} To address the limitations of the Q-only Estimator, IPW introduces the concept of propensity scores, \(g(x)\), estimating the probability of treatment given covariates. This method reweights observations to balance covariate distributions between treated and untreated groups. However, IPW is sensitive to extreme propensity scores (\(\hat{g}(x)\) close to 0 or 1 ), which can lead to unstable estimates.

\textbf{Augmented IPW (AIPW):} AIPW enhances IPW by combining the propensity score with outcome modeling. This dual adjustment stabilizes the estimation process, particularly when extreme propensity scores are present, offering a more robust approach to treatment effect estimation.

\subsection{CausalBERT Parameter Detail}
\label{sec:causalbert_param}
The CausalBERT model, similar to DistilBERT, utilizes a model dimension of 768 with an input maximum length of 512 tokens. The embeddings layer consists of word embeddings with a size of \([30\,522, 768]\) and position embeddings with a size of \([512, 768]\). The transformer layers, comprising six blocks, have attention and feedforward networks, each with parameter vectors of size \([768, 768]\) and \([768, 3072]\) respectively. For the downstream tasks, such as treatment assignment prediction and outcome regression, the model uses fully connected layers with a hidden dimension of 200. In total, the model contains approximately 66 million parameters, with key components distributed as follows: embeddings layer (24 million), transformer layers (43 million), and downstream layers (0.3 million).

\subsection{Propensity score distribution}
\label{sec:propensity_score_balance}
To investigate why temperature scaling most significantly enhances IPW and AIPW at higher confounder strengths (as shown in Figure~\ref{fig:combined} (Left) and Figure~\ref{fig:temp_boost_aipw}), we begin by defining the overlap metric as follows:
\[
\text{Overlap} = \frac{1}{N} \left( \sum_{i: T_i = 0} \hat{g_i} + \sum_{i: T_i = 1} (1 - \hat{g_i}) \right),
\]
where $N$ is the total number of observations, $\hat{g_i}$ is the predicted propensity score for observation $i$, and $T_i$ is the treatment label for observation $i$. This \text{Overlap} metric assesses the similarity of propensity score distributions between treatment/control groups. A higher \text{Overlap} indicates a better balance of observed covariates between groups, which enhances the reliability of causal inference. 

\begin{figure}[t]
  \includegraphics[width=\columnwidth]{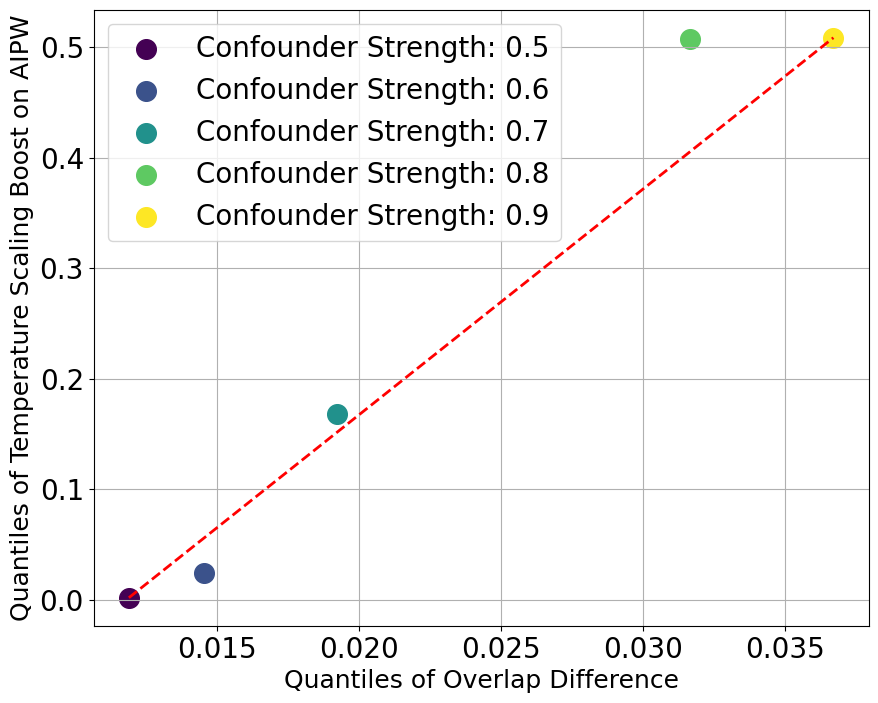}
  \caption{Q–Q Plot of Overlap Differences (the overlap 
increment after temperature scaling) vs. Temperature Scaling Boost on AIPW (the AIPW error ratio reduction after the temperature scaling), with $\alpha$=0.4.}
  \label{fig:overlap_AIPWboost}
\end{figure}

Figure~\ref{fig:overlap_AIPWboost} shows the Q–Q Plot of Overlap Differences (the overlap 
increment after temperature scaling) vs. Temperature Scaling Boost on AIPW across different confounder strengths $p$ from 0.9 to 0.5 and true ATE $u=-0.4$, computed from an overall validation set in cross-validation with CausalBERT(\(\alpha\)=0.4). We observe that at higher confounder strengths, particularly \( p = 0.9 \), there is a marked increase in the overlap increment, which correlates with significant improvements in AIPW estimation accuracy and vice versa for lower confounder strength. This trend suggests that the temperature scaling interventions effectively mitigate the distortions in the propensity score distributions that are more pronounced at higher confounder levels. Consequently, the propensity score estimate is rectified towards a more moderate range, preventing the extreme values that typically skew the analysis at high confounder strengths. This improved covariate balance directly enhances AIPW estimation, demonstrating the benefits of scaling interventions in strongly confounded scenarios.

To enhance the robustness of our findings in Figure~\ref{fig:overlap_AIPWboost}, we extend our analysis by examining the Pearson Correlation Coefficients \cite{schober2018correlation} across multiple true ATE values ($u = -0.4, -0.35, -0.3$). This involves two specific correlations: first, between the confounder strength and the increment in overlap, and second, between the overlap increment and the boost in AIPW estimation accuracy. Across these different true ATE scenarios, we compute and average the Pearson correlation coefficients, with the first correlation yielding a value of \(0.974 \pm 0.022\), and the second yielding a value of \(0.503 \pm 0.061\). Both correlations are consistently positive, indicating that a larger confounder strength is always associated with a more pronounced increase in overlap after temperature scaling, and that a larger increase in overlap is associated with a more significant boost in AIPW estimation accuracy. This consistent positivity affirms the observations from the Q–Q Plot and underscores the efficacy of temperature scaling in adjusting for confounder-related distortions within propensity score distributions, particularly in settings with high confounder strength.

\subsection{Keywords for School-Related Topics}
\label{sec:keywords_list}

The following lists detail the keywords used to identify mentions of various school-related topics in reviews, adapted from the work by \cite{harris2022}. Refer to Table~\ref{tab:keywords} for the specific keywords categorized by topic.

\begin{table*}[!ht]
\centering
\caption{Keywords used to identify school-related topics in reviews.}
\label{tab:keywords}
\begin{tabularx}{\textwidth}{|p{2.1cm}|X|}
\hline
\textbf{Topic} & \textbf{Keywords List} \\
\hline
Bullying & bully, bullying, harassment, intimidation, teasing, taunting, tormenting, victimization, abuse, threatening, coercion, humiliation, exclusion, cyberbullying, aggression, peer pressure, verbal abuse, physical abuse, emotional abuse, marginalization, ostracization, discrimination, hazing, stalking \\
\hline
Administration & administrator, board, counselor, counselors, coaches, handle, governance, head, advisor, superintendent, headmaster, vice, administration, policymaker, dean, policies, policy, staff, policymakers, coach, director, faculty, educators, educator, admin, assistant, headmistress, professional, administrative, leadership, principle, principal, administrators, principals, management \\
\hline
Academic Performance & exams, grade, score, standards, benchmark, ratings, evaluations, exam, reputation, performance, review, achievement, assessment, tests, rated, results, test scores, scores, grades, evaluation, star, rate, standard, academic, test, performing, benchmarks, assessments, success, rating \\
\hline
Extracurricular & athletic, club, extracurricular, music, teams, basketball, team, band, sports, drama, art, football, clubs, dance, athletics, soccer, choir, orchestra, volleyball, cheerleading, theater, debate, speech, track, swimming, tennis, robotics, student council, volunteer \\
\hline
Curriculum & language, reading, spanish, books, projects, courses, pe, writing, course, arts, science, assignments, book, subjects, math, english, social, history, lessons, write, homework, curricular, ap, curriculum, subject, physical \\
\hline
\end{tabularx}
\end{table*}

\subsection{Data Process Details}
\label{sec:data_collection_detail}
\subsubsection{Selected Period Define}
To determine the most relevant time period for analyzing the influence of specific topics in school reviews, we employ a systematic approach to identify the four consecutive years with the highest frequency of topic mentions across schools. This method involves aggregating the review data annually for each school and isolating those reviews that contain keywords associated with the topic of interest. We then count the number of unique schools discussing these topics each year. By examining these annual counts, we pinpoint the time span where the conversation around each topic is most concentrated. This focused analysis ensures that our causal inference study is grounded in the period of maximal relevance for each educational aspect under consideration. 

For each topic, once we define the time period, we then aggregate the review data for each school in this period to get school-level data. The data is further refined by excluding schools with fewer than five reviews and ensuring that concatenated non-keyword comments comprise at least 100 tokens. In this process, we define the treatment, outcome, and covariate for each school accordingly, as we discuss in Section~\ref{sec:Dataset}. The summary of the school-level data is shown in Appendix \ref{app:school_stats}.

\subsubsection{Bootstrap Detail}
For the real-world data experiment, to ensure statistical robustness, we utilize a bootstrap sampling method without replacement, selecting equal numbers of treated and control schools for each topic to create balanced datasets. Specifically, based on the school-level data summary shown in Appendix \ref{app:school_stats}, we sample 9,000 schools for `bullying,' 3,200 schools for `academic performance,' 2,000 schools for `administration,' 2,600 schools for `extracurricular,' and 4,000 schools for `curriculum.' This process is repeated six independent times for each topic to mitigate any sampling bias and provide a comprehensive overview of the causal impacts. The experimental results are then averaged across these six iterations to present a consolidated finding on how specific aspects influence school ratings. 

\subsubsection{Reviews Concatenate}
When concatenating non-keyword reviews for each school, we insert the ‘[SEP]’ special token between individual reviews. This token is preserved during training to serve as a delimiter, indicating the transition between separate reviews from the same school. To accommodate the DistilBERT model's token limit of 512, we employ a two-step truncation method for concatenated reviews exceeding 450 tokens. Initially, up to four sentences are included in each component review unless adding another sentence would exceed 450 tokens. If the token limit has not been reached after this initial pass, additional sentences are sequentially added from each review in rounds, continuing until reaching the token limit or exhausting all sentences. This approach is based on the assumption that the first several sentences of a review typically contain the most relevant information. The aim is to balance the depth of content detail within each review and the breadth of including multiple reviews, ensuring comprehensive coverage without exceeding token constraints.

\subsubsection{Confounder Insert}
In our semi-synthetic data experiments, confounder text is randomly inserted at different positions within the reviews, specifically at locations marked by `[SEP]` tokens, at the beginning or at the end of the review texts. This insertion ensures that the confounder text is separated from the existing content by a `[SEP]` token, maintaining the structure of the original reviews. This approach simulates the addition of an authentic single review, seamlessly integrating the confounder text to reflect realistic review scenarios while preserving the integrity of individual reviews.

\subsection{Semi-Synthetic Data Detail}
\label{sec:semi-synthetic_Data_appendix}

In this section, we introduce the probability distribution framework we employ for the semi-synthetic data. Building on the definitions in Section~\ref{sec:estimatingframework}, we expand our probability model to include a confounder class for each subject $i$, represented as $(X_i, Y_i, T_i, C_i)$. Here, $C_i \in \{1, 2\}$ categorizes the confounder. 

Outcome distributions are modeled to assess the effect of treatment across confounder classes. When $T=1$ and $C=1$, outcomes follow a target distribution $Y \sim \mathcal{N}(u_2, 0.3)$, contrasting with $Y \sim \mathcal{N}(u_1, 0.3)$ when $T=0$ and $C=1$. For Class 2, irrespective of the treatment, outcomes are both modeled as $Y \sim \mathcal{N}(u_2, 0.3)$, indicating uniform effects in the absence of confounding text. In this example, the conditional ATE for Class 1 is $(u_2-u_1)$, and for Class 2 is $0$, then the true ATE $u = (u_2-u_1)/2$ for the whole dataset. 

The probability of confounding class assignment is evenly distributed with $P(C=1) = P(C=2) = 0.5$. Class 1 subjects receive academic challenge posts inserted into their observed text, whereas Class 2 subjects' text remains unaltered, simulating environments with varying levels of textual confounding. Within these classes, the treatment assignment probabilities are designed to reflect their respective confounding impacts: $P(T=1 | C=1) = 1-p$ and $P(T=0 | C=1) = p$, contrasting with $P(T=1 | C=2) = p$ and $P(T=0 | C=2) = 1-p$. The magnitude of \( p \) directly modulates the strength of the confounding effect in our model. As \( p \) approaches 1, it signifies a strong correlation between the treatment assignment \( T \) and the confounder class \( C \), indicating robust confounding. Conversely, when \( p \) nears 0.5, treatment assignment becomes effectively random within each class \( C \), implying minimal or no confounding influence. Thus, \( p \) serves as a key parameter to adjust the intensity of confounding in our study. In our case, we vary p from 0.9 to 0.5.

The selection of \(u_1\) and \(u_2\) in our experiment is informed by the statistical properties of our 13,361 population dataset. The primary objective is to ensure that our sample not only includes a sufficient number of instances but also reflects a high sampling quality that aligns closely with the target distribution. Specifically, we set \(u_2 = -0.3\), as the median and mean of \(Y | T=1\) in our dataset hover around -0.3. For \(u_1\), we explore values within the range [-0.2, -0.1, 0.0, 0.1, 0.2, 0.3, 0.4, 0.5], and ultimately select 0.3, 0.4, and 0.5, where the resulting sample distribution closely matches our target. Ultimately, we choose a sample size of 5,000 for our analysis, ensuring both robustness and relevance in our evaluation of CausalBERT’s performance.

\subsection{Additional Results}
\label{additional_results}
\begin{figure}[t]
  \includegraphics[width=0.98\linewidth]{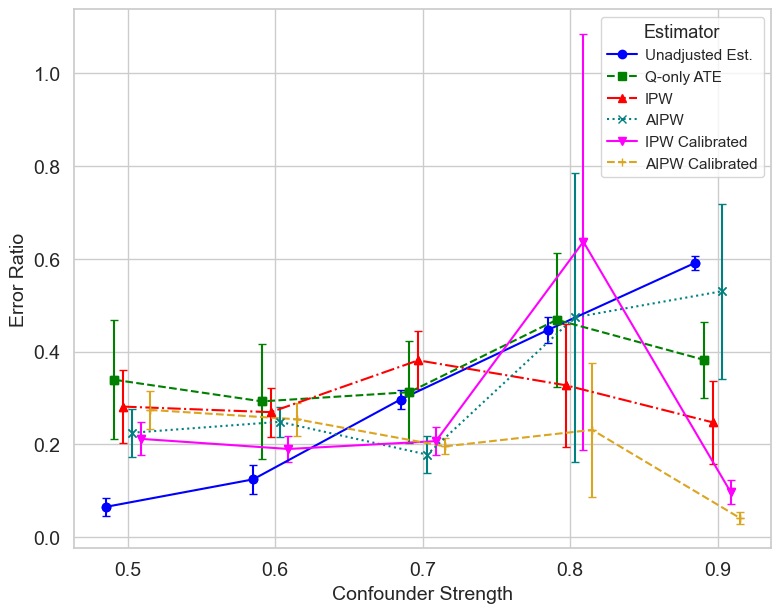}
  \caption {Estimator performance at a true ATE of $u=-0.35$, consistent with the settings in Figure~\ref{fig:ate_est}.}
  \label{fig:ate35}
\end{figure}

\begin{figure}[t]
  \includegraphics[width=0.98\linewidth]{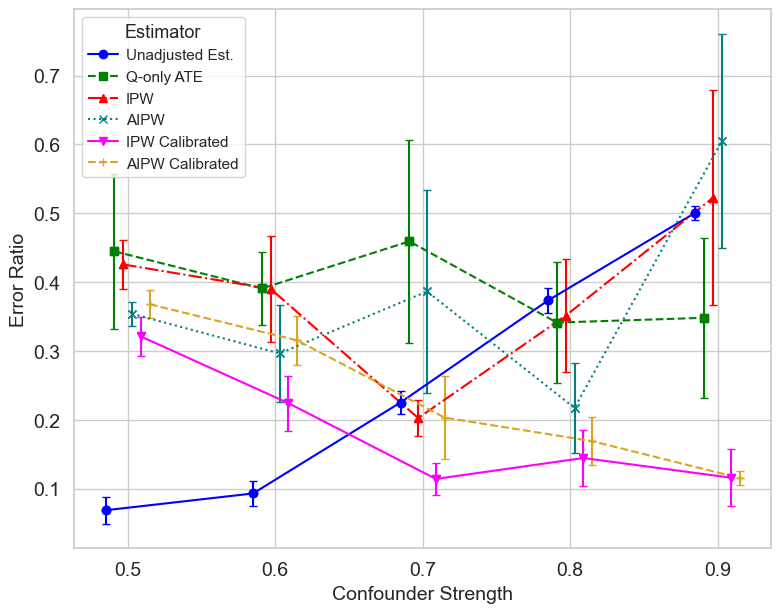}
  \caption {Estimator performance at a true ATE of $u=-0.4$, consistent with the settings in Figure~\ref{fig:ate_est}.}
  \label{fig:ate40}
\end{figure}

\paragraph{Estimation for others true ATE}
Figures~\ref{fig:ate35} and \ref{fig:ate40} show the performance of different estimators across various confounder strengths for semi-synthetic datasets where the true ATEs are -0.35 and -0.4, respectively.

\begin{figure}[ht]
  \includegraphics[width=0.98\linewidth]{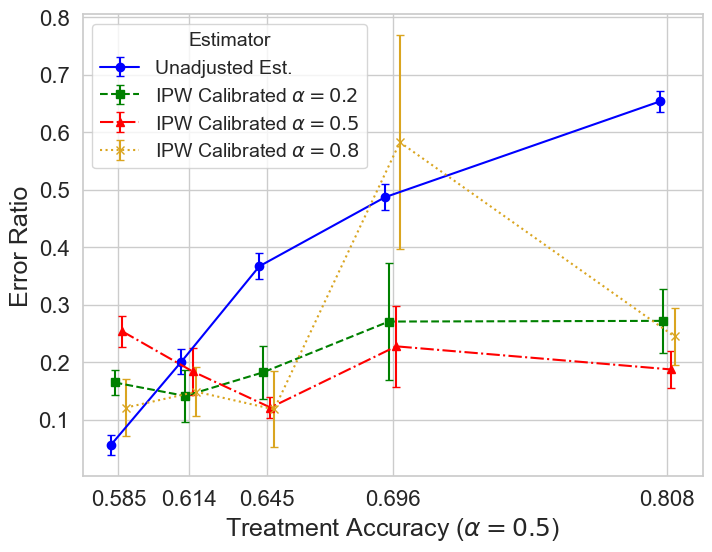} \hfill
  \caption {Error ratio comparison for CausalBERT under varying \(\alpha\) values (0.2, 0.5, 0.8) indicated by treatment accuracy (trained with $\alpha = 0.5$) on a semi-synthetic dataset with a fixed true ATE $u=-0.3$, for the IPW Calibrated model.}
  \label{fig:g_weight_comparisons_ipw}
\end{figure}

\paragraph{Effect of $\alpha$}
Figure~\ref{fig:g_weight_comparisons_ipw} shows the effect of $\alpha$ on the IPW Calibrated model, mirroring the results found on AIPW in Figure~\ref{fig:combined} (Right).

\begin{figure}[t]
  \includegraphics[width=0.98\linewidth]{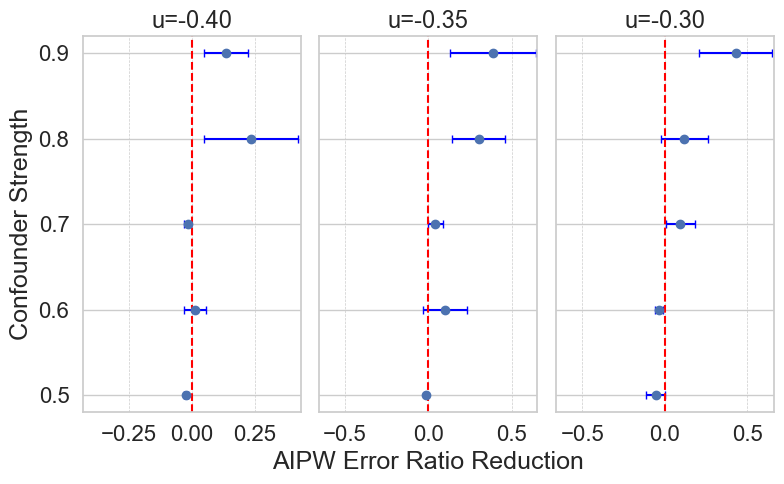}
  \caption {Error ratio decrease by temperature scaling on AIPW estimations}
  \label{fig:temp_boost_aipw}
\end{figure}

\paragraph{Temperature Scaling}
Figure~\ref{fig:temp_boost_aipw} shows the average error reduction provided by temperature scaling on the AIPW ATE estimates, mirroring the results found on IPW in Figure~\ref{fig:combined} (Left).

\paragraph{Integrated Gradients} 

\begin{figure}[t]
  \includegraphics[width=\columnwidth]{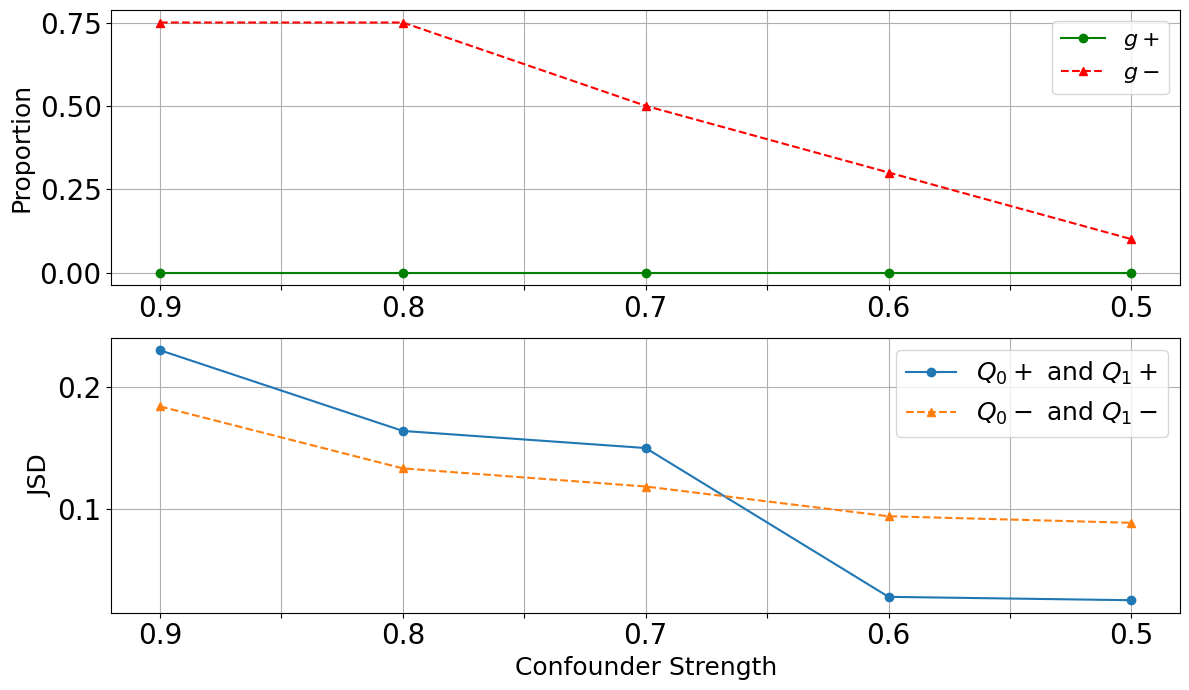}
  \caption{Application of Integrated Gradients on CausalBERT models trained across semi-synthetic datasets with a fixed true ATE \(u=-0.3\) across a varying of confounder strengths from 0.9 to 0.5. The analysis identifies the top 20 tokens that influence the model’s predictions across different output components. The top graph displays the proportion of these tokens from `$g+$` and `$g-$` originating from the inserted confounder template, and the bottom graph compares the Jensen–Shannon divergence(JSD) between the weighted tokens from `$Q_{0}^+$` and `$Q_{1}^+$` as well as `$Q_{0}^-$` and `$Q_{1}^-$`. Both graphs are presented across varying confounder strengths.}
  \label{fig:ig_syn}
\end{figure}

In this section, we extend our qualitative analysis on semi-synthetic datasets with a fixed true ATE \(u=-0.3\) and varying confounder strengths from 0.9 to 0.5 using Integrated Gradients. Figure~\ref{fig:ig_syn} reveals how top tokens(top 20) from `$g^+$`, `$g^-$`, `$Q_{0}^+$`, `$Q_{1}^+$`, `$Q_{0}^-$` and `$Q_{1}^-$` respond to changes in confounder strengths within CausalBERT. The top graph tracks the proportion of tokens from `$g^+$` and `$g^-$` that derive from the confounder template, showing how the model's reliance on confounder-driven features for treatment prediction varies with confounder strength. The bottom graph employs Jensen–Shannon divergence(JSD)\footnote{The Jensen–Shannon divergence (JSD) is a symmetric measure of the similarity between two probability distributions \cite{MENENDEZ1997307}}  to measure the similarity between weighted tokens from `$Q_{0}^+$` and `$Q_{1}^+$` as well as `$Q_{0}^-$` and `$Q_{1}^-$`. 

We observe from the top graph that the proportion of words in `$g^-$` that originate from the confounder template decreases linearly as confounder strength is reduced from 0.9 to 0.5. This aligns with our expectations from the semi-synthetic dataset design. Under strong confounder conditions, a substantial portion of the non-treatment text incorporates inserted confounder text, making these features strong predictors of non-treatment. As confounder strength weakens, and the distribution of confounder text between treatment and non-treatment groups becomes more balanced, the model's reliance on these features for predicting treatment diminishes accordingly.

In the bottom graph, we observe a decrease in the JSD between the weighted token lists $Q_{0}^+$ and $Q_{1}^+$, as well as between $Q_{0}^-$ and $Q_{1}^-$, as confounder strength diminishes. This trend indicates that not only are the same terms present in both lists, but their relative weights also converge, suggesting a decreasing distinction in the outcome prediction signals for treated and untreated groups with weaker confounder influence. This result aligns with the design of our semi-synthetic data, which expects diminishing differences in reviews between treated and control groups as confounder strength decreases. Additionally, the range of change in JSD is notably greater for $Q_{0}^+$ and $Q_{1}^+$ than for $Q_{0}^-$ and $Q_{1}^-$. This pattern aligns with the structural design of our semi-synthetic data, where a higher target mean outcome ($u_1=0.3$) is associated with the presence of inserted confounder text, in contrast to a lower target mean ($u_2=-0.3$), which is mostly derived from Class 2 (without inserted confounder text). This design implies that the inserted confounder text serves as an indicator of higher overall ratings. Thus, the prediction signals for increasing outcomes exhibit greater sensitivity to shifts in confounder strength, as reflected by the larger divergence observed in $Q_{0}^+$ and $Q_{1}^+$. These findings collectively highlight CausalBERT’s nuanced ability to detect, adapt to, and accurately represent subtle changes and biases in treatment effects, confirming its effectiveness in evaluating the impacts on outcomes under varying conditions of confounder strength.

\begin{figure}[t]
  \includegraphics[width=\columnwidth]{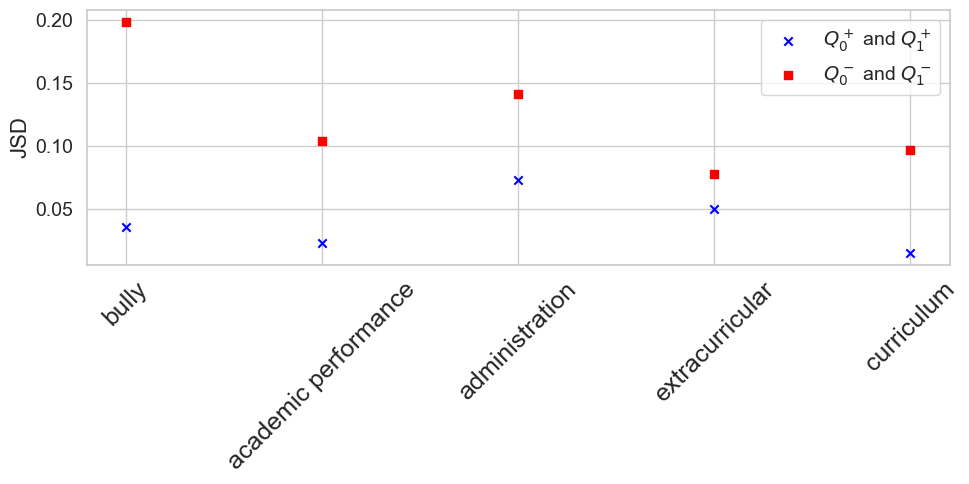}
  \caption{Comparison of the Jensen–Shannon divergence(JSD) between the weighted tokens from `$Q_{0}^+$` and `$Q_{1}^+$` as well as `$Q_{0}^-$` and `$Q_{1}^-$` across all topics. All weighted tokens are derived from the Integrated Gradients analysis applied to each topic from a single bootstrap sample.}
  \label{fig:jsd_real}
\end{figure}

\paragraph{Weighted Token Comparison}
In this section, we focus on the Integrated Gradients analysis applied to real data, emphasizing the weighted token differences between top terms (top 20) \(Q_1^+\) and \(Q_0^+\), as well as \(Q_1^-\) and \(Q_0^-\). Figure~\ref{fig:jsd_real} shows the Jensen–Shannon divergence(JSD) between the weighted tokens from `$Q_{0}^+$` and `$Q_{1}^+$` as well as `$Q_{0}^-$` and `$Q_{1}^-$` across all topics. we observe a distinct pattern across all topics: the JSD for the outcome-decreasing predictions top terms (`$Q_{0}^-$` and `$Q_{1}^-$`) consistently exhibits greater divergence compared to the outcome-increasing predictions top terms(`$Q_{0}^+$` and `$Q_{1}^+$`), particularly notable in the 'bully' topic. This trend suggests that, compared to positive comments, negative comments about schools often cover a broader and more varied range of concerns, based on specific school attributes.

\begin{figure*}[ht]
  \centering
  \includegraphics[width=\textwidth]{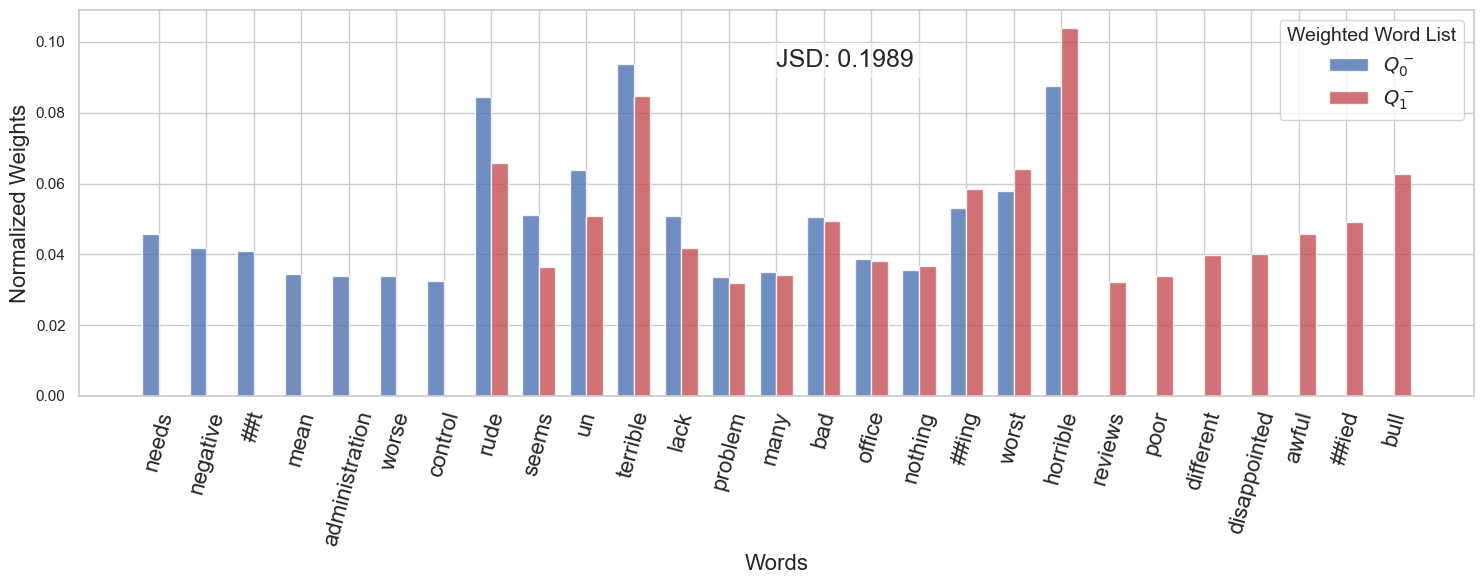}
  \caption{The weight distribution for top 20 weighted tokens from `$Q_{0}^-$` and `$Q_{1}^-$` for 'bullying' topic. We also include the JSD between these two weighted tokens.}
  \label{fig:Q-_bully}
\end{figure*}

\begin{figure*}[ht]
  \centering
  \includegraphics[width=\textwidth]{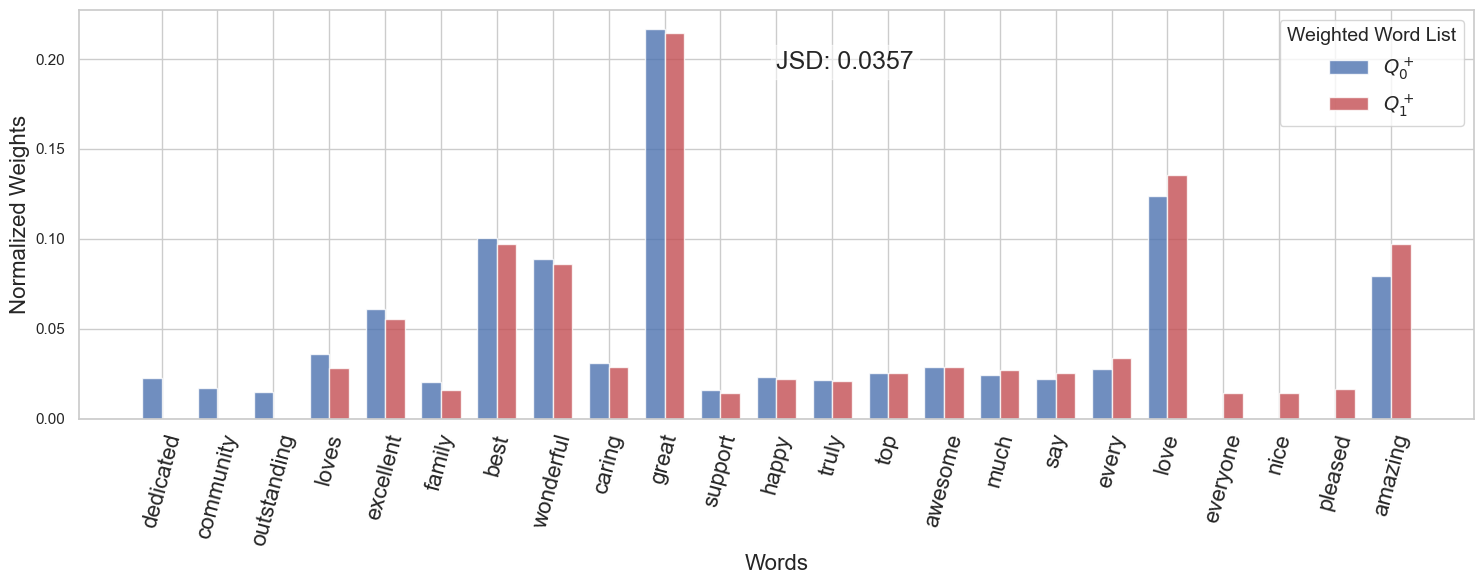}
  \caption{The weight distribution for top 20 weighted tokens from `$Q_{0}^+$` and `$Q_{1}^+$` for 'bullying' topic with JSD between these weighted tokens.}
  \label{fig:Q+_bully}
\end{figure*}

\begin{figure*}[ht]
  \centering
  \includegraphics[width=\textwidth]{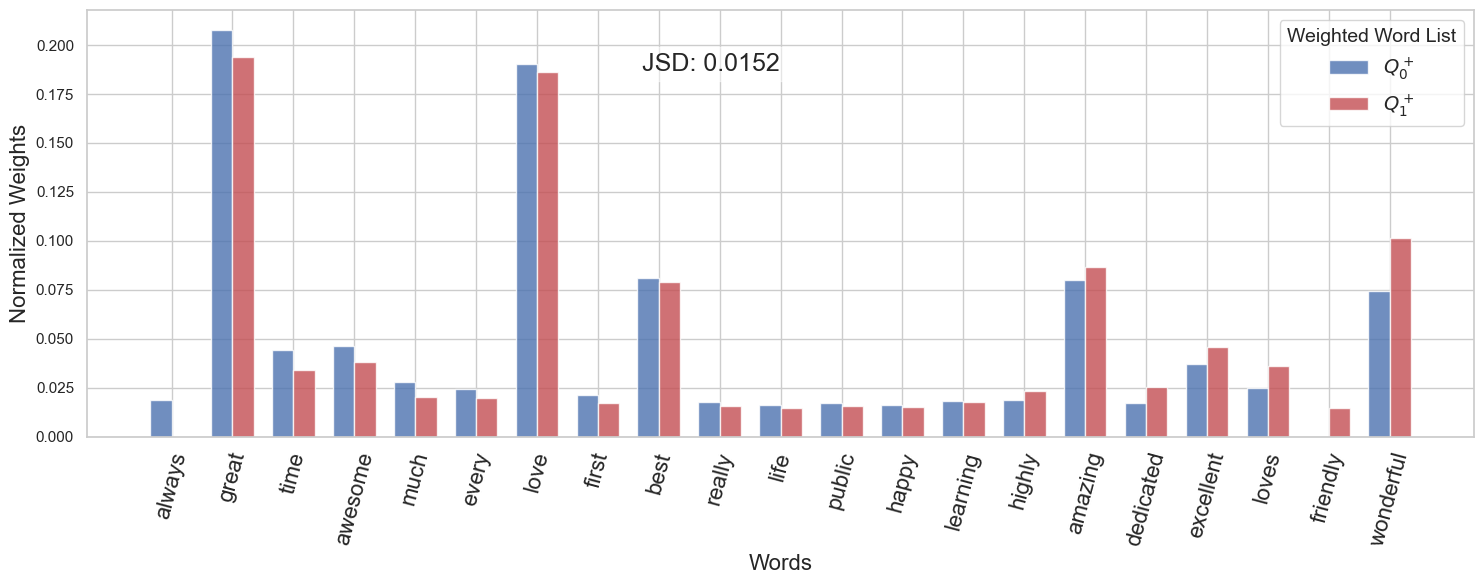}
  \caption{The weight distribution for top 20 weighted tokens from `$Q_{0}^+$` and `$Q_{1}^+$` for 'curriculum' topic with JSD between these weighted tokens.}
  \label{fig:Q+_curriculum}
\end{figure*}

\begin{figure*}[ht]
  \centering
  \includegraphics[width=\textwidth]{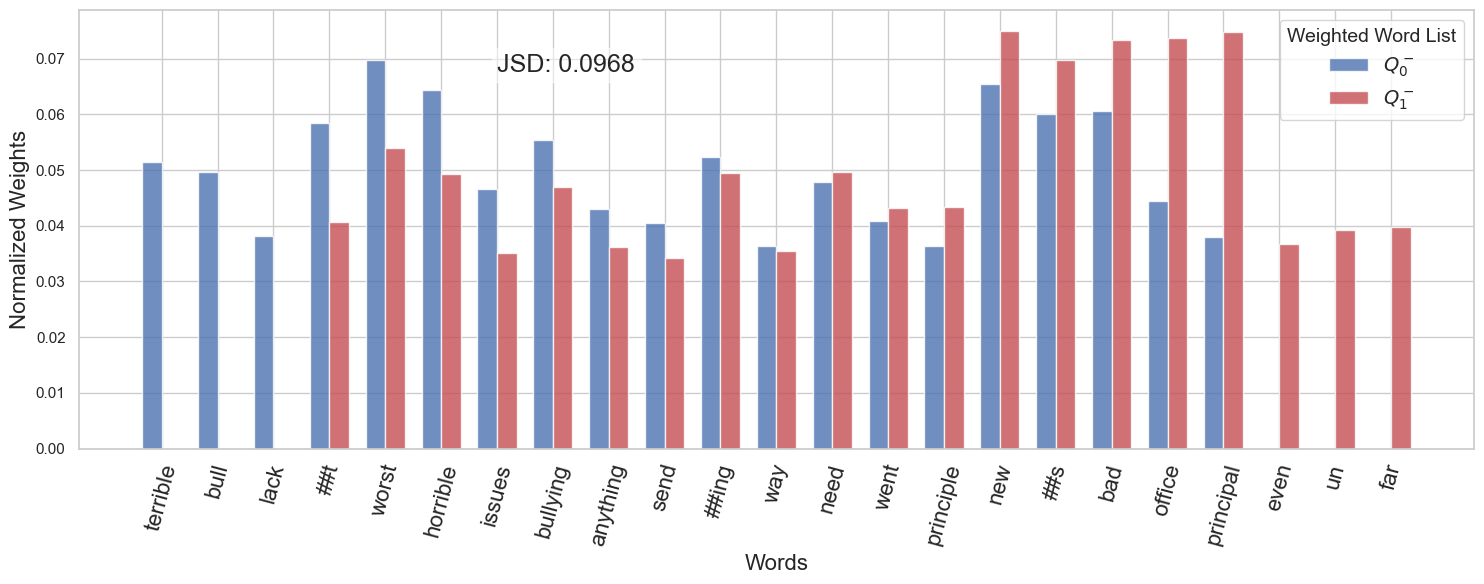}
  \caption{The weight distribution for top 20 weighted tokens from `$Q_{0}^-$` and `$Q_{1}^-$` for 'curriculum' topic with JSD between these weighted tokens.}
  \label{fig:Q-_curriculum}
\end{figure*}

To provide additional context for Figure~\ref{fig:jsd_real}, we present the detailed weight distribution of the top terms used in the analysis along with their corresponding JSD values for several topics ('bullying' and 'curriculum'), as shown in Figures~\ref{fig:Q-_bully}, \ref{fig:Q+_bully}, \ref{fig:Q+_curriculum}, and \ref{fig:Q-_curriculum}. Generally, the observed variations between control and treatment groups in outcome predictions are subtle, especially for outcome-increasing predictions. Universally positive terms such as 'great,' 'love,' and 'amazing' are prevalent in both groups (Figures~\ref{fig:Q+_bully}\&\ref{fig:Q+_curriculum}), indicating a consensus on the attributes that positively impact school perceptions. For outcome-decreasing predictions, while negatively charged terms like 'bad' and 'horrible' are common in both scenarios (Figure~\ref{fig:Q-_curriculum}), there remain slight distinctions, highlighting subtler variations in concerns that negatively impact school ratings. For example, in the 'bullying' topic (Figure~\ref{fig:Q-_bully}), we notice that the term 'administration' is predominantly featured in the control group's negative predictors, highlighting that in the absence of bullying, issues like poor administration significantly impact school ratings. Conversely, this term is less pronounced in the treated group, suggesting that the presence of bullying tends to overshadow other administrative problems in influencing school ratings.

\end{document}